\documentclass[11pt,letterpaper]{article}
\PassOptionsToPackage{breaklinks}{hyperref}
\pdfoutput=1  

\usepackage{paclic33}
\usepackage{times}
\usepackage{latexsym}
\usepackage{amssymb}  

\usepackage{natbib}
\usepackage{url}
\usepackage{paralist}
\usepackage{relsize}
\usepackage{fancyvrb}
\usepackage{hyperref}
\usepackage[capitalise]{cleveref}
\usepackage{multirow}
\usepackage{tikz}
\usetikzlibrary{arrows,positioning}

\usepackage[rounded,nounderscore]{syntax}
\newcommand\form[1]{\textit{#1}}
\newcommand\amr[1]{\texttt{\smaller #1}}
\newcommand\Amr[1]{\texttt{#1}}  
\newcommand\mc[3]{\multicolumn{#1}{#2}{#3}}
\newcommand\mr[2]{\multirow{#1}{*}{#2}}
\newcommand\cm{\checkmark}

\title{AMR Normalization for Fairer Evaluation}

\author{Michael Wayne Goodman \\
  Nanyang Technological University \\
  Singapore \\
  {\tt goodmami@uw.edu}
}
\date{}

\begin{document}
\maketitle

\begin{abstract}
  Abstract Meaning Representation \citep[AMR;][]{banarescu-EtAl:2013:LAW7-ID} encodes the meaning of sentences as a directed graph and Smatch \citep{cai2013smatch} is the primary metric for evaluating AMR graphs.
  Smatch, however, is unaware of some meaning-equivalent variations in graph structure allowed by the AMR Specification and gives different scores for AMRs exhibiting these variations.
  In this paper I propose four normalization methods for helping to ensure that conceptually equivalent AMRs are evaluated as equivalent.
  Equivalent AMRs with and without normalization can look quite different---comparing a gold corpus to itself with relation reification alone yields a difference of 25 Smatch points, suggesting that the outputs of two systems may not be directly comparable without normalization.
  The algorithms described in this paper are implemented on top of an existing open-source Python toolkit for AMR and will be released under the same license.
\end{abstract}

\section{Introduction}

Abstract Meaning Representation \citep[AMR;][]{banarescu-EtAl:2013:LAW7-ID} encodes the meaning of sentences in a rooted, directed acyclic graph of concepts (labeled nodes) and relations (labeled edges).
It was introduced as being to semantics what the Penn Treebank \citep{Marcus:1994:PTA:1075812.1075835} was to syntax---a simple pairing of sentences and hand-authored annotations---and aimed to coalesce multiple aspects of semantic annotation that had previously been done separately, such as named entity recognition, role labeling, and coreference resolution, into one form.

Research efforts targeting AMR often use the Smatch metric \citep{cai2013smatch} for evaluation.
Smatch views AMR graphs as bags of triples and attempts to find a mapping of nodes between two AMRs that results in the highest F-score in terms of matching triples.
The result is a single score for a list of AMR pairs.
As AMR encodes many aspects of meaning in one graph, some have found it useful to divide up the parts of the graph that Smatch evaluates so as to inspect a parser's aptitude in each task \citep{damonte-etal-2017-incremental}.
Nevertheless, Smatch remains the primary underlying method for comparing AMRs and thus ensuring that it is a fair metric is important for the task of semantic parsing.

The AMR Specification\footnote{\url{https://github.com/amrisi/amr-guidelines}} describes some features of the representation that expand its expressiveness and improve its legibility, such as reifying graph edges to nodes so that the meaning of the edge can be used by other parts of the graph, and rules for inverting edges so the graph can be linearized into the PENMAN format \citep{matthiessen1991text}.
The specification says that these alternations express the same meaning, but they result in different triples used by Smatch for comparison.

In this paper, I investigate the effects these differences have on comparison and propose normalization methods to aid in resolving them.
Normalization is intended as a preprocessing step to evaluation and is done to both the gold and test corpus.
The purpose is not to yield higher Smatch scores or to change system outputs, but to ensure that conceptually equivalent AMRs evaluate as equivalent and that no system is unfairly penalized or rewarded.

\section{Background}
\label{sec:bg}

While AMR and its PENMAN notation are often considered one and the same, I find that distinguishing them aids the discussion of the Smatch metric, so in this section I explain all three in turn.

\subsection{PENMAN Graph Notation}
\label{sec:bg:penman}

PENMAN notation for AMR is a variation of Sentence Plan Language \citep{kasper1989spl} for the PENMAN project \citep{matthiessen1991text}.
The notation is applicable to graphs that are:
\begin{inparaenum}[(1)]
\item directed and acyclic (DAGs),
\item connected,
\item with a distinguished root called the \form{top}, and
\item with labeled nodes and edges.%
\end{inparaenum}%
\footnote{No technical reason precludes cyclic and unlabeled graphs in PENMAN but I will consider these errors for this paper.}
The basic syntax for nodes and edges is as follows:

\begin{grammar}
  <node> ::= `(' <id> `/' <node-label> <edge>* `)'

  <edge> ::= `:'<edge-label> (<const>|<id>|<node>)
\end{grammar}

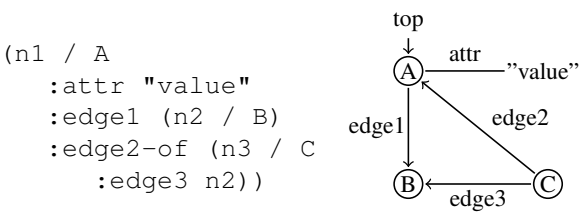
\begin{figure}[ht]
  \centering\small
  \begin{BVerbatim}
(n1 / A
   :attr "value"
   :edge1 (n2 / B)
   :edge2-of (n3 / C
      :edge3 n2))
\end{BVerbatim}
  \hspace{1em}
  \begin{tikzpicture}[baseline=-14ex,semithick,every node/.style={inner sep=0pt,outer sep=1pt,minimum size=1.2em}]
    \footnotesize
    \node[inner sep=1pt] (top) {top};
    \node[draw, circle] (a) [below=6pt of top] {A};
    \node[draw, circle] (b) [below=30pt of a] {B};
    \node[draw, circle] (c) [right=40pt of b] {C};
    \node (a-attr) [right=30pt of a] {"value"};
    \path[draw] (top) edge [->] (a);
    \path[draw] (a) edge[-,edge label={attr},pos=.5,sloped] (a-attr);
    \path[draw] (a) edge [->,edge label={edge1},swap] (b);
    \path[draw] (c) edge [->,edge label={edge2},pos=.4,swap] (a);
    \path[draw] (c) edge [->] node [below] {edge3} (b);
  \end{tikzpicture}
  \caption{PENMAN notation and the equivalent graph}
  \label{fig:penman-graph}
\end{figure}

The recursion of nodes as targets of edges can only capture projective structures such as trees.
In order to encode multiple roots (besides the top node), edges are inverted so the source becomes the target by appending \amr{-of} to the edge label.
For reentrancies, node identifiers, also and hereafter called \form{variables}, are reused.\footnote{\citet{kasper1989spl} allowed node attributes and edges to be distributed across multiple references to the node but I will not consider this feature in this paper.}
\Cref{fig:penman-graph} shows an example PENMAN serialization, with all the above features, along with the graph it describes.

This paper uses the relative terms \form{parent} and \form{child} for the nodes of an edge in the tree structure and \form{source} and \form{target} for nodes in the directed graph edges (i.e., such that parent$=$source in regular edges and parent$=$target in inverted edges).
Edges whose target is a constant are \form{attributes}.
The place where a node specifies its label is the \form{node definition}.

AMR, described in the next section, uses PENMAN notation to serialize its graph structure.
While AMR and PENMAN share a history, the graph notation is not restricted to AMR and could in principle be used for any graphs that meet its criteria.
For example it has also been used to encode Dependency Minimal Recursion Semantics \citep[DMRS;][]{copestake-2009-invited} for neural text generation \citep{Hajdik:et-al:2019} and machine translation \citep{Goodman:2018}.

\subsection{Abstract Meaning Representation}
\label{sec:bg:amr}

Where PENMAN notation is the serialization format, Abstract Meaning Representation \citep{banarescu-EtAl:2013:LAW7-ID} is the semantic framework.
As AMR graphs encode semantic information, it refers to node labels as \form{concepts}, to edges as \form{relations}, and to edge labels as \form{roles}.
AMR defines in the specification and annotation documentation\footnote{\url{https://www.isi.edu/~ulf/amr/lib/roles.html}} the inventories of valid concepts and roles and their usage.
An AMR graph serialized in PENMAN notation, as in \cref{fig:drive1-penman}, is simply called an \form{AMR}, but it can also be represented as a sequence of triples, as in \cref{fig:drive1-triples}.
Node labels are represented by \amr{instance} triples\footnote{Some prefer \amr{instance-of} but the choice is arbitrary; I use \amr{instance} to avoid the ramifications of inverted edges.} and the top node is indicated with the \amr{:TOP} triple.

\begin{figure}[ht]
  \centering\small
  \begin{BVerbatim}
(d / drive-01
   :ARG0 (h / he)
   :manner (c / care-04
      :polarity -))
\end{BVerbatim}
  \caption{AMR for \form{He drives carelessly.}}
  \label{fig:drive1-penman}
\end{figure}

\begin{figure}[ht]
  \centering\small
\begin{BVerbatim}
instance(d, drive-01) ^
instance(h, he) ^
instance(c, care-04) ^
TOP(top, d) ^
ARG0(d, h) ^
manner(d, c) ^
polarity(c, -)
\end{BVerbatim}
  \caption{Triples for \form{He drives carelessly.}}
  \label{fig:drive1-triples}
\end{figure}

Several PENMAN graphs may correspond to the same set of triples.
A tree-structured graph as in \cref{fig:drive1-penman} has limited options---the branches for \amr{:ARG0} and \amr{:manner} can swap positions, but that's it---but graphs with reentrancies can ``rotate'' on the reentrant nodes.
For example, the graph in \cref{fig:penman-graph} could also be represented as in \cref{fig:penman-graph-rotated} or 26 other ways.\footnote{There are 6 rotations and each rotation has 2 or 6 arrangements by swapping branch positions; more are possible when the top node is not fixed.}
These alternative serializations do not affect the meaning as determined by the triples (used in evaluation as discussed below), but they can cause issues for systems that learn the serialized character sequences \citep[e.g.,][]{konstas-etal-2017-neural,van2017neural}.
\citet{konstas-etal-2017-neural} found that human annotators preferred to insert non-core and inverted relations in the same order as in the original sentence, which leaked ordering information.

\begin{figure}[ht]
  \centering\small
  \begin{BVerbatim}
(n1 / A
   :edge1 (n2 / B
      :edge3-of (n3 / C
         :edge2 n1))
   :attr "value")
\end{BVerbatim}
  \caption{Alternative serialization of the graph in \cref{fig:penman-graph}}
  \label{fig:penman-graph-rotated}
\end{figure}


While AMR lacks a notion of scope and has no direct model theoretic interpretation,\footnote{\citet{bos2016expressive} proposed a transformation to first-order logic and also found that a minor change to AMR could allow negation scope to be accurately encoded. \citet{stabler2017reforming} extended this work and included tense and number features.} it can encode partial scope information implicitly.
For example, the AMRs for \form{the fast car is red} and \form{the red car is fast} would differ only by which concept, \amr{fast-02} or \amr{red-02}, is the top of the graph (AMR calls this ``focus'').
If the examples were, instead, \form{the fast car that is red} and \form{the red car that is fast}, then \amr{car} would be the top of both and the triples would be the same, but the PENMAN serializations could differ.
Furthermore, reentrancies in AMR present a choice of which occurrence of a variable gets the node definition.
It would not be surprising, therefore, for annotators to prefer different PENMAN arrangements for sentences with the same triples, as in \cref{fig:amr-dog1,fig:amr-dog2}.
Put another way, the PENMAN serialization can encode information not present in the triples.

\begin{figure}[ht]
  \centering\small
  \begin{BVerbatim}
(b / bite-01
   :ARG0 (d / dog
      :ARG0-of (c / chase-01
         :ARG1 (b2 / boy)))
   :ARG1 b2)
\end{BVerbatim}
  \caption{AMR for \form{The dog chasing the boy bit him.}}
  \label{fig:amr-dog1}
\end{figure}

\begin{figure}[ht]
  \centering\small
  \begin{BVerbatim}
(b / bite-01
   :ARG0 (d / dog)
   :ARG1 (b2 / boy
      :ARG1-of (c / chase-01
         :ARG0 d)))
\end{BVerbatim}
  \caption{AMR for \form{The dog bit the boy whom it chased.}}
  \label{fig:amr-dog2}
\end{figure}

The AMR Specification also describes equivalent\footnote{Equivalent only by the AMR Specification, not necessarily logical equivalence by a mapping of AMR to logical forms.} variants where the triples do in fact differ.
One case is the roles \amr{:domain} and \amr{:mod}, which are considered equivalent in the inverse (i.e., \amr{:domain-of} is equivalent to \amr{:mod}, etc.).
The other case is reified relations, where a relation between two nodes becomes a binary node, which is useful when the relation itself interacts with other parts of the graph.
These are explained further in \cref{sec:norm:inv,sec:norm:re}.

\subsection{Smatch}
\label{sec:bg:smatch}

Smatch \citep{cai2013smatch} is the primary metric used for AMR evaluation.
It estimates the ``overlap'' between two AMRs by finding a mapping of variables that optimizes the number of matching triples.
Precision is defined as $\frac{M}{T}$ and recall as $\frac{M}{G}$ where $M$ is the number of matching triples, $T$ is the number of test triples, and $G$ is the number of gold triples,\footnote{The Smatch utility I use (see \cref{sec:setup}) does not specify gold and test, only the first and second arguments. Swapping these arguments swaps precision and recall. I set the gold corpus to the second argument.}  and the final Smatch score is the F-score of these two.
Finding an ideal mapping is an NP-complete task, so Smatch approximates it using greedy search with random restarts to avoid local optima.
As regular and inverted relations in AMR are the same when presented as triples, any rearrangement of the PENMAN form for the same triples (as discussed in \cref{sec:bg:amr}) will yield the same results as long as the top node does not change, exempting search errors.

Smatch is na\"ive with respect to AMR-specific interpretations of PENMAN graphs---it only considers the most direct translation of PENMAN graphs to triples.
It does not consider equivalent alternations where the triples do change (such as \amr{:domain} vs \amr{:mod} alternations and relation reifications) as equivalent, and these alternations will lead to score differences.
Smatch is also not robust to subtly invalid graphs, such as inverted edges whose source (i.e., child in the tree structure) is a constant.\footnote{Only nodes, not constants, may specify relations. These invalid graphs occur occasionally in the output of some parsers.}
In this case, the triple will be ignored completely, leading to an inflated score.

Moreover, Smatch gives no credit for a correct role or value unless both are correct.
For example, the first line in the Little Prince corpus is \form{Chapter 7} with the AMR \amr{(c / chapter :mod 7)}, but all three parsers I tested failed to output the correct relation (one gave \amr{:quant 7}, another \amr{:li 7}, and another \amr{:op1 7}).
They are therefore all penalized in recall for missing the \amr{:mod 7} relation and again in precision for their incorrect attempt, and none get credit for the correct value of \amr{7}.
Omitting the relation entirely (e.g., \amr{(c / chapter)}) yields a higher score, but that's hardly ideal.

The AMR normalizations described in this paper ensure equivalent AMRs have the same triples and thus the same score.
In addition, two of the normalizations involve reification which replaces a single triple with several, and this presents a tradeoff: it can allow ``partial credit'' for getting the role or the value correct, but getting both wrong hurts the score worse than getting a single relation wrong.

\section{AMR Normalization}

This section describes two meaning-preserving AMR normalizations and two meaning-augmenting normalizations.
The first two include canonical role inversions and relation reification, while the latter two include attribute reification and PENMAN structure preservation.

\subsection{Canonical Role Inversions}
\label{sec:norm:inv}

The roles of inverted relations are marked with an \amr{-of} suffix, and generally they are deinverted by removing the suffix.
AMR, however, specifies several roles whose canonical form contains the suffix \amr{-of}, namely \amr{:consist-of}, \amr{:prep-on-behalf-of}, and \amr{:prep-out-of}, and the inverse form of these therefore requires an additional suffix (e.g., \amr{:prep-out-of-of}).
In addition there is \amr{:mod} which is equivalent to the inverse of \amr{:domain}, and vice-versa.\footnote{The specification suggests that \amr{:mod} \emph{is} the inverse of \amr{:domain}, but that could not be true as \amr{:mod} appears in attribute relations and a relation's source cannot be a constant.}
If a gold corpus contained \amr{:mod} while the test corpus used \amr{:domain-of}, Smatch would not see these as equivalent and the score would drop.

By normalizing inverted roles to their canonical forms, such as \amr{:domain-of} $\rightarrow$ \amr{:mod}, \amr{:consist} $\rightarrow$ \amr{:consist-of-of}, the Smatch score will not differ for such alternations.
Some may argue that normalizing invalid roles such as \amr{:consist} in this way is meaning-altering, but as the na\"ive inversions of these roles are not separately defined roles in AMR there is no chance of conflation, and in this case I take the position that practicality beats purity.

\subsection{Relation Reifications}
\label{sec:norm:re}

Some specific relations in AMR can be reified into concepts with separate relations for the original relation's source and target.
For example, \cref{fig:drive2-penman} is equivalent to \cref{fig:drive1-penman} with \amr{:manner} reified to \amr{have-manner-91}.
While its possible to reify every eligible relation, in practice all are collapsed unless it is necessary to have the node, so \cref{fig:drive1-penman} would generally be preferred over \cref{fig:drive2-penman}.

\begin{figure}[hpt]
  \centering\small
  \begin{BVerbatim}
(d / drive-01
   :ARG0 (h / he)
   :ARG1-of (h2 / have-manner-91
      :ARG2 (c / care-04
               :polarity -)))
\end{BVerbatim}
  \caption{AMR for \form{He drives carelessly} with \amr{:manner} reified to \amr{have-manner-91}}
  \label{fig:drive2-penman}
\end{figure}

\begin{figure}[ht]
  \centering\small
\begin{BVerbatim}
(d / drive-01
   :ARG0 (h / he)
   :ARG1-of (h2 / have-manner-91
      :ARG2 (c / care-04)
      :polarity -))
\end{BVerbatim}
  \caption{AMR for \form{He doesn't drive carefully.}}
  \label{fig:drive3-penman}
\end{figure}

There are three situations where reification is useful:
\begin{inparaenum}[(1)]
\item when the meaning of the relation itself is the focus or the argument of another concept instance;
\item when it breaks a cycle in the graph; and
\item when an annotator uses a ``shortcut'' role in a relation.
\end{inparaenum}
Situation (1) is the only case that is strictly necessary.
For example, \cref{fig:drive3-penman} is used to express \form{He doesn't drive carefully}, where the have-manner property is negated rather than the manner itself.
The breaking of cycles in situation (2) is possible because reification replaces an edge with a node and two outgoing edges, thus becoming a new root (but not necessarily the graph's top).
These kinds of reifications ensure that the graph remains a DAG---a property that may be useful for some applications.
The ``shortcut'' roles of situation (3) are a feature of the AMR Editor \citep{hermjakob2013amr} provided as a convenience to annotators.
They are always reified automatically by the editor and therefore might be considered not part of the official role inventory in the AMR framework.
Annotators not using the editor, however, might use them as they are listed in the specification, so it is still useful to reify these in normalization.

In implementation, reification is not complicated.
The process uses a defined mapping of roles to AMR fragments containing the reified concept and the roles that capture the original relation's source and target.
A sample of these definitions is shown in \cref{tab:reif-list}; the full list is given in \cref{sec:app1}.
Reification uses this mapping to replace some relation \amr{(a :<role> b)} with \amr{(a :<source>-of (c / <concept> :<target> b))} for regular relations and \amr{(a :<target>-of (c / <concept> :<source> b))} for inverted relations.
Reification used in normalization will always have one inverted edge as the original AMR would not have had any way to focus the pre-reified relation.

\begin{table}
  \small
  \begin{tabular}{llll}
    Role               & Concept                & Source      & Target      \\\hline
    \amr{:degree}      & \amr{have-degree-92}   & \amr{:ARG1}  & \amr{:ARG2}  \\
    \amr{:manner}      & \amr{have-manner-91}   & \amr{:ARG1}  & \amr{:ARG2}  \\
    \amr{:purpose}     & \amr{have-purpose-91}  & \amr{:ARG1}  & \amr{:ARG2}  \\
  \end{tabular}
  \caption{Sample of reification definitions}
  \label{tab:reif-list}
\end{table}

Collapsing, or dereifying, nodes to edges is slightly more complicated because there are more restrictions on when it can be applied.
A node can only be collapsed if it does not participate in relations (including the \amr{:TOP} relation) other than those resulting from reification.\footnote{While it is possible to pull out and collapse the information relating to the reified relation and leave in place the node and its additional relations, I do not do so here.}
For example, \amr{have-manner-91} in \cref{fig:drive2-penman} can be collapsed but it cannot be in \cref{fig:drive3-penman} because in the latter it is involved in the \amr{:polarity} relation.
The change to the graph itself is just the opposite of reification: \amr{(a :<source>-of (c / <concept> :<target> b))} becomes \amr{(a :<role> b)} and \amr{(a :<target>-of (c / <concept> :<source> b)} becomes \amr{(a :<role>-of b)}.

There are additional complexities when the reification mapping is not one-to-one; that is, when it maps multiple relations to the same concept or a single relation to multiple alternative concepts.
For the first case, normalization always introduces a new node for each reified relation, even when multiple relations on the same node are mapped to the same concept.
This case only occurs with the shortcut roles \amr{:employed-by}/\amr{:role} and \amr{:subset}/\amr{:superset}.
For the second case the relations will not be reified because it is undecidable which of the competing concepts should be used, and likewise in dereification information would be lost by collapsing both concepts to the same relation.
This case occurs with \amr{:poss} reifying to either \amr{own-01} or \amr{have-03}, and \amr{:beneficiary} reifying to either \amr{benefit-01} and \amr{receive-01}.

The effect of reification on the Smatch score can be large.
By reifying one relation to a node with two relations, the net total of triples increases by two.
In the gold corpus (see \cref{sec:setup,sec:analysis}), roughly 15\% of triples were reifiable, so a fully-reified corpus would contain roughly 30\% more triples.
The result is that Smatch will require more time and memory to compute a score, and the search for the variable mapping may become less stable because there are more nodes to search over.
This normalization can affect the Smatch score by amplifying certain kinds of errors and giving partial credit for others.
\Cref{tab:re-smatch-ex} shows a gold item (the top AMR for \form{five apples}) and several test AMRs with various differences.
The Collapsed column shows the Smatch score between the gold and test AMRs when the relations are left as-is, and the Reified column shows the score when both gold and test are reified.
Smatch's preference for missing versus incorrect relations becomes a dispreference unless the test AMR's role differs and is not reifiable (\amr{:unit} in \cref{tab:re-smatch-ex}).

\begin{table}[h]
\begin{tabular}{lrr}
 AMR                              & Collapsed & Reified \\\hline
 \amr{(a / apple :quant 5)}       & -         & - \\\hline
 \amr{(a / apple)}                & 0.80      & 0.57 \\
 \amr{(a / apple :quant {\bf 1})} & 0.67      & 0.80 \\
 \amr{(a / apple {\bf :mod} 5)}   & 0.67      & 0.80 \\
 \amr{(a / apple {\bf :mod 1})}   & 0.67      & 0.60 \\
 \amr{(a / apple {\bf :unit} 5)}  & 0.67      & 0.50 \\
 \amr{(a / apple {\bf :unit 1})}  & 0.67      & 0.50 \\
\end{tabular}
\caption{Difference in Smatch score with and without reification (top is gold, rest are test, bold are differences)}
\label{tab:re-smatch-ex}
\end{table}

\subsection{Attribute Reification}
\label{sec:norm:attr}

As mentioned in \cref{sec:bg:smatch}, Smatch silently drops triples whose source is not a valid variable, leading to inflated scores.
While canonical role inversions (such as \amr{:domain-of} to \amr{:mod}) help here, the situation can be completely averted by reifying every constant into a node with a new unique variable and with the constant as the node's concept.
For example, \amr{:mod 7} becomes \amr{:mod (\_ / 7)}.
The result is not meaning-equivalent as the alternation is not provided by the AMR Specification, but it will at least allow each triple to be considered in evaluation.
The effect on Smatch is that each attribute triple is replaced with a relation and a concept triple, thus increasing the number of triples by one for each constant.
It also allows for partial credit, similar to reification.

\subsection{PENMAN Structure Preservation}
\label{sec:norm:struct}

\Cref{sec:bg:amr} described two kinds of variation in PENMAN that correspond to the same triples: the order of serialized relations on a node and which occurrence of a node contains the node definition.
As discussed, these differences can be used to encode nuance or hints to the surface form that the AMR annotates.
In order to preserve the information encoded by the location of node definitions, additional \amr{:TOP} relations may be used to indicate which node is the top of the node being defined.
These parallel the tree structure rather than the DAG, so they do not invert if the child of an inverted relation (i.e., the relation's source) is a node definition.\footnote{These \amr{:TOP} relations could lead to a cyclic structure so it is not recommended as a general annotation practice.}
Inserting these relations into an AMR with $n$ nodes results in $n-1$ new triples as one is not inserted for the top node in the graph.
The effect on Smatch is a boost in the score of AMRs that define nodes in the same place.

\section{Experiment Setup}
\label{sec:setup}

For information about roles and their reifications I use the AMR 1.2.6 Specification\footnote{\url{https://github.com/amrisi/amr-guidelines}} and the annotator documentation of roles as of May 1, 2019.\footnote{\url{https://www.isi.edu/~ulf/amr/lib/roles.html}}
For reification I use all non-ambiguous mappings, which excludes \amr{:beneficiary} and \amr{:poss}, and for dereification I also exclude mappings of shortcut roles.
My experiments use the training portion of the freely-available Little Prince corpus (version 1.6).\footnote{\url{https://amr.isi.edu/download.html}}
For reading and writing PENMAN graphs I use the open-source Penman package (version 0.6.2) for Python.\footnote{\url{https://github.com/goodmami/penman/}}
I used JAMR \citep{flanigan2016cmu},\footnote{Semeval-2016 branch as of March 21, 2019:\\\url{https://github.com/jflanigan/jamr}} CAMR \citep{wang-EtAl:2016:SemEval},\footnote{Master branch as of February 19, 2018:\\\url{https://github.com/c-amr/camr}} and AMREager \citep{damonte-etal-2017-incremental}\footnote{Master branch as of April 11, 2019:\\\url{https://github.com/mdtux89/amr-eager}} for producing system outputs.
All systems use their included models trained on the LDC2015E86 (SemEval Task 8) data, which is out-of-domain for the Little Prince corpus but the parsers then all use comparable models.
For comparison I use Smatch \citep{cai2013smatch}.\footnote{Master branch as of August 21, 2018: \url{https://github.com/snowblink14/smatch/}}

\section{Corpus Analysis}
\label{sec:analysis}

I first inspect the corpus to understand the distribution of normalizable AMRs.
Table \ref{tab:size} shows the number of nodes and triples in The Little Prince corpus (1,274 AMRs) for both gold annotations and system outputs.
These counts are used for calculating the percentages in Tables \ref{tab:roleinv} and \ref{tab:reco}.

\begin{table}[ht]
  \centering\small
  \begin{tabular}{lrr}
    Corpus       & \# Nodes & \# Triples \\\hline
    Gold         & 8,189    & 16,832       \\
    JAMR         & 8,115    & 15,509       \\
    CAMR         & 7,404    & 13,922       \\
    AMREager     & 7,461    & 15,226       \\
  \end{tabular}
  \caption{Corpus sizes}
  \label{tab:size}
\end{table}

Table \ref{tab:roleinv} shows the percentage of graphs and triples that have the non-canonical \amr{:domain-of} and \amr{:mod-of} relations.
They do not appear in the gold annotations, CAMR output, or AMREager output, but do in the JAMR output along with a small number of \amr{:consist} relations (not shown), and no corpus used non-canonical inversions of the \amr{:prep-*} relations.
This is not unexpected, as the gold corpus does not contain any instances of these roles, so data-driven parsers would have no examples to learn from.
A parser that assembles the graph and inverts as necessary to serialize may be susceptible.

\begin{table}[ht]
  \centering\small
  \begin{tabular}{lrrrr}
                 & \mc{2}{c}{\% \Amr{:domain-of}} & \mc{2}{c}{\% \Amr{:mod-of}} \\

    Corpus       & Graphs & Triples & Graphs & Triples        \\\hline
    Gold         & 0      & 0    & 0      & 0           \\
    JAMR         & 5.81   & 0.52 & 8.63   & 0.80        \\
    CAMR         & 0      & 0    & 0      & 0           \\
    AMREager     & 0      & 0    & 0      & 0           \\
  \end{tabular}
  \caption{Non-canonical role inversions}
  \label{tab:roleinv}
\end{table}

Table \ref{tab:reco} shows the percentage of graphs and relations that are reifiable and the percentage of graphs and nodes that are collapsible.
All systems have roughly as many reifiable graphs and relations as the gold corpus.
CAMR is the only system that outputs reified relations that can be collapsed, although the number is miniscule.

\begin{table}[ht]
  \centering\small
  \begin{tabular}{lrrrr}
                 & \mc{2}{c}{\% Reifiable} & \mc{2}{c}{\% Collapsible} \\

    Corpus       & Graphs & Rels  & Graphs & Nodes    \\\hline
    Gold         & 78.96  & 15.23 & 0      & 0        \\
    JAMR         & 73.94  & 13.07 & 0      & 0        \\
    CAMR         & 68.68  & 14.22 & 0.16   & 0.03     \\
    AMREager     & 76.92  & 17.02 & 0      & 0        \\
  \end{tabular}
  \caption{Reifiable relations and collapsible nodes}
  \label{tab:reco}
\end{table}

Using Smatch to compare two versions of the gold corpus---one original and one with reified relations---yields an F-Score of $0.75$, or a drop of 25 Smatch points.
This result is an estimate of the range of score variation when a system perfectly reproduces the gold corpus but makes the opposite decision regarding reification.

\section{System Evaluation}

Here I test the effect the normalizations have on Smatch when evaluating system outputs to the gold corpus.
\Cref{tab:smatch} shows the results of the three systems with various normalizations.
While JAMR was the only parser that output non-canonical roles, normalizing the roles did not help its score; in fact, the score dropped slightly.
Some of JAMR's non-canonical roles were inverted relations to constants, so Smatch was ignoring them.
Normalizing them would thus hurt the score unless the normalized relations were correct.
Reification (both kinds) generally led to higher scores, meaning that most relations that were reified were fully or partially correct.
One result that stands out is structure preservation; for both JAMR and AMREager it led to decreased scores but it helped CAMR, showing that CAMR is more likely to place node definitions where an annotator would.
Finally, the normalization helped AMREager close the gap with JAMR, and in some configurations even surpass it.

\begin{table}[ht]
  \centering\small
  \begin{tabular}{lccccrrr}
           & \mc{4}{c}{Normalization} & \mc{3}{c}{Score} \\

    System           & I   & A   & R   & S   & P             & R             & F             \\\hline
    \mr{9}{JAMR}     &     &     &     &     & 0.60          & 0.56          & 0.58          \\
                     & \cm &     &     &     & 0.60          & 0.55          & 0.57          \\
                     &     & \cm &     &     & 0.61          & 0.56          & 0.58          \\
                     &     &     & \cm &     & 0.63          & 0.57          & 0.60          \\
                     &     &     &     & \cm & 0.59          & 0.55          & 0.57          \\
                     & \cm &     & \cm &     & 0.63          & 0.57          & 0.60          \\
                     &     & \cm & \cm &     & 0.64          & 0.57          & 0.60          \\
                     & \cm & \cm & \cm &     & 0.64          & 0.57          & 0.60          \\
                     & \cm & \cm & \cm & \cm & 0.61          & 0.56          & 0.59          \\\hline
    \mr{9}{CAMR}     &     &     &     &     & 0.67          & 0.56          & 0.61          \\
                     & \cm &     &     &     & 0.67          & 0.56          & 0.61          \\
                     &     & \cm &     &     & 0.67          & 0.55          & 0.60          \\
                     &     &     & \cm &     & \textbf{0.70} & 0.57          & \textbf{0.63} \\
                     &     &     &     & \cm & 0.68          & \textbf{0.58} & \textbf{0.63} \\
                     & \cm &     & \cm &     & 0.69          & 0.57          & \textbf{0.63} \\
                     &     & \cm & \cm &     & \textbf{0.70} & 0.56          & 0.62          \\
                     & \cm & \cm & \cm &     & \textbf{0.70} & 0.56          & 0.62          \\
                     & \cm & \cm & \cm & \cm & \textbf{0.70} & \textbf{0.58} & \textbf{0.63} \\\hline
    \mr{9}{AMREager} &     &     &     &     & 0.57          & 0.52          & 0.55          \\
                     & \cm &     &     &     & 0.57          & 0.52          & 0.55          \\
                     &     & \cm &     &     & 0.57          & 0.53          & 0.55          \\
                     &     &     & \cm &     & 0.61          & 0.57          & 0.59          \\
                     &     &     &     & \cm & 0.59          & 0.54          & 0.56          \\
                     & \cm &     & \cm &     & 0.61          & 0.57          & 0.59          \\
                     &     & \cm & \cm &     & 0.60          & \textbf{0.58} & 0.59          \\
                     & \cm & \cm & \cm &     & 0.60          & \textbf{0.58} & 0.59          \\
                     & \cm & \cm & \cm & \cm & 0.61          & 0.57          & 0.59          \\
  \end{tabular}
  \caption{Smatch results comparing gold to system outputs with the original graphs, canonical role inversions (I), attribute reification (A), relation reification (R), and structure preservation (S)}
  \label{tab:smatch}
\end{table}

\section{Related Work}

\citet{konstas-etal-2017-neural} normalized AMRs is a destructive way in order to reduce data sparsity for their character-based neural parser and generator.
My normalization methods can also reduce sparsity but they also generally increase the size and complexity of the graph, so it's not clear if it would aid character-based models.
\citet{damonte-etal-2017-incremental} found that parsers do well on different sub-tasks, such role labeling and word-sense disambiguation, and ran Smatch on different subsets of the triples in order to highlight a parser's performance in each task.
In addition, \citeauthor{damonte-etal-2017-incremental} also found that Smatch weighted certain error types more than others, although they looked at more application-specific error types, like the representation of proper names.
In contrast, I compare using the full graphs as the goal is normalization, not specialization.
My normalization methods are mostly compatible with the subtask evaluation of \citealt{damonte-etal-2017-incremental} but some the evaluation tasks look for certain roles which disappear on reification.
\citet{anchieta2019sema} also noticed that Smatch gives more weight to the top node of the graph, but they reached different conclusions.
Where I proposed adding \amr{:TOP} relations to all nodes to preserve the PENMAN structure, they discard the \amr{:TOP} node, meaning that the AMRs for \form{the fast car is red} and \form{the red car is fast} are evaluated as equivalent.
\citet{barzdins-gosko-2016-riga} presented extensions to Smatch including a visualization of per-sentence error patterns and an ensemble selection from multiple test AMRs per gold AMR.
The latter extension could in principle be combined with the normalization procedures I have described, however it would need to be augmented to allow for the normalizations of the gold corpus as well as the test corpus.

\section{Conclusion and Future Work}
\label{sec:conc}

AMR provides flexibility with the way that equivalent graphs are encoded.
This flexibility can make life easier for annotators and parsers alike, but it also means that evaluation tools not aware of these allowed alternations can give unfair results.
I introduced four normalization methods in this paper.
Of these, canonical role inversion, relation reification, and attribute reification are intended to tame the variation that can reasonably appear in parser outputs.
The fourth, PENMAN structure preservation, makes evaluation more strictly account for annotation choices which may implicitly encode subtle distinctions in meaning, like scope or nuance.

The evaluation results when comparing a normalized test corpus to the similarly normalized gold corpus are not drastically different.
I think this result is a good thing, particularly because comparing a corpus to itself with and without normalization has a very large difference in scores.
It suggests that normalization, done to both sides, resolves small differences.
While one parser I tested, CAMR, maintained its lead with normalized outputs, the third-place parser AMREager nearly caught up to the second-place JAMR.
The relative changes in evaluation scores may important for determining state-of-the-art parsers or for shared task competitions.

The normalizations may be useful not only for evaluation but for preprocessing for data-driven workflows.
By removing sources of variation, data sparsity can be reduced which could benefit parser training.
The increase in graph size due to the normalization, however, may counteract the benefits.
I leave this question open to future research.

The code for this paper is available online at \url{https://github.com/goodmami/norman}.

\section*{Acknowledgments}

Thanks to three anonymous reviewers, to Francis Bond, and to Guy
Lapalme for their comments, suggestions, and corrections.

\bibliography{paclic33}
\bibliographystyle{acl_natbib}

\appendix
\section{Relation Reifications}
\label{sec:app1}

\begin{table}[hptb]
  \small
  \begin{tabular}{lllllll}
    Role               & Concept                   & Source      & Target      & Reifies & Dereifies & Shortcut \\\hline
    \amr{:accompanier} & \amr{accompany-01}        & \amr{:ARG0} & \amr{:ARG1} & \cm     & \cm       &          \\
    \amr{:age}         & \amr{age-01}              & \amr{:ARG1} & \amr{:ARG2} & \cm     & \cm       &          \\
    \amr{:beneficiary} & \amr{benefit-01}          & \amr{:ARG0} & \amr{:ARG1} &         &           &          \\
    \amr{:beneficiary} & \amr{receive-01}          & \amr{:ARG2} & \amr{:ARG0} &         &           &          \\
    \amr{:cause}       & \amr{cause-01}            & \amr{:ARG1} & \amr{:ARG0} & \cm     &           & \cm      \\
    \amr{:concession}  & \amr{have-concession-91}  & \amr{:ARG1} & \amr{:ARG2} & \cm     & \cm       &          \\
    \amr{:condition}   & \amr{have-condition-91}   & \amr{:ARG1} & \amr{:ARG2} & \cm     & \cm       &          \\
    \amr{:cost}        & \amr{cost-01}             & \amr{:ARG1} & \amr{:ARG2} & \cm     &           & \cm      \\
    \amr{:degree}      & \amr{have-degree-92}      & \amr{:ARG1} & \amr{:ARG2} & \cm     & \cm       &          \\
    \amr{:destination} & \amr{be-destined-for-91}  & \amr{:ARG1} & \amr{:ARG2} & \cm     & \cm       &          \\
    \amr{:domain}      & \amr{have-mod-91}         & \amr{:ARG2} & \amr{:ARG1} & \cm     & \cm       &          \\
    \amr{:duration}    & \amr{last-01}             & \amr{:ARG1} & \amr{:ARG2} & \cm     & \cm       &          \\
    \amr{:employed-by} & \amr{have-org-role-91}    & \amr{:ARG0} & \amr{:ARG1} & \cm     &           & \cm      \\
    \amr{:example}     & \amr{exemplify-01}        & \amr{:ARG0} & \amr{:ARG1} & \cm     & \cm       &          \\
    \amr{:extent}      & \amr{have-extent-91}      & \amr{:ARG1} & \amr{:ARG2} & \cm     & \cm       &          \\
    \amr{:frequency}   & \amr{have-frequency-91}   & \amr{:ARG1} & \amr{:ARG2} & \cm     & \cm       &          \\
    \amr{:instrument}  & \amr{have-instrument-91}  & \amr{:ARG1} & \amr{:ARG2} & \cm     & \cm       &          \\
    \amr{:li}          & \amr{have-li-91}          & \amr{:ARG1} & \amr{:ARG2} & \cm     & \cm       &          \\
    \amr{:location}    & \amr{be-located-at-91}    & \amr{:ARG1} & \amr{:ARG2} & \cm     & \cm       &          \\
    \amr{:manner}      & \amr{have-manner-91}      & \amr{:ARG1} & \amr{:ARG2} & \cm     & \cm       &          \\
    \amr{:meaning}     & \amr{mean-01}             & \amr{:ARG1} & \amr{:ARG2} & \cm     &           & \cm      \\
    \amr{:mod}         & \amr{have-mod-91}         & \amr{:ARG1} & \amr{:ARG2} & \cm     & \cm       &          \\
    \amr{:name}        & \amr{have-name-91}        & \amr{:ARG1} & \amr{:ARG2} & \cm     & \cm       &          \\
    \amr{:ord}         & \amr{have-ord-91}         & \amr{:ARG1} & \amr{:ARG2} & \cm     & \cm       &          \\
    \amr{:part}        & \amr{have-part-91}        & \amr{:ARG1} & \amr{:ARG2} & \cm     & \cm       &          \\
    \amr{:polarity}    & \amr{have-polarity-91}    & \amr{:ARG1} & \amr{:ARG2} & \cm     & \cm       &          \\
    \amr{:poss}        & \amr{own-01}              & \amr{:ARG0} & \amr{:ARG1} &         &           &          \\
    \amr{:poss}        & \amr{have-03}             & \amr{:ARG0} & \amr{:ARG1} &         &           &          \\
    \amr{:purpose}     & \amr{have-purpose-91}     & \amr{:ARG1} & \amr{:ARG2} & \cm     & \cm       &          \\
    \amr{:quant}       & \amr{have-quant-91}       & \amr{:ARG1} & \amr{:ARG2} & \cm     & \cm       &          \\
    \amr{:role}        & \amr{have-org-role-91}    & \amr{:ARG0} & \amr{:ARG2} & \cm     &           & \cm      \\
    \amr{:source}      & \amr{be-from-91}          & \amr{:ARG1} & \amr{:ARG2} & \cm     & \cm       &          \\
    \amr{:subevent}    & \amr{have-subevent-91}    & \amr{:ARG1} & \amr{:ARG2} & \cm     & \cm       &          \\
    \amr{:subset}      & \amr{include-91}          & \amr{:ARG2} & \amr{:ARG1} & \cm     &           & \cm      \\
    \amr{:superset}    & \amr{include-91}          & \amr{:ARG1} & \amr{:ARG2} & \cm     &           & \cm      \\
    \amr{:time}        & \amr{be-temporally-at-91} & \amr{:ARG1} & \amr{:ARG2} & \cm     & \cm       &          \\
    \amr{:topic}       & \amr{concern-02}          & \amr{:ARG0} & \amr{:ARG1} & \cm     & \cm       &          \\
    \amr{:value}       & \amr{have-value-91}       & \amr{:ARG1} & \amr{:ARG2} & \cm     & \cm       &          \\
  \end{tabular}
  \caption{Full mapping of roles and concepts used for reification, dereification, and editor shortcuts}
  \label{tab:full-reif-list}
\end{table}

\end{document}